\documentclass{article}

\usepackage{PRIMEarxiv}

\usepackage[utf8]{inputenc} 
\usepackage[T1]{fontenc}    
\usepackage{hyperref}       
\usepackage{url}            
\usepackage{booktabs}       
\usepackage{amsfonts}       
\usepackage{nicefrac}       
\usepackage{microtype}      
\usepackage{lipsum}
\usepackage{authblk}
\usepackage{fancyhdr}       
\usepackage{graphicx}       
\graphicspath{{media/}}     
\usepackage{times}
\usepackage{soul}
\usepackage{multirow}
\usepackage{url}
\usepackage{amsmath}
\usepackage{amsthm}
\usepackage{booktabs}
\usepackage{array}
\usepackage{algorithm}
\usepackage{algorithmic}
\usepackage[switch]{lineno}
\usepackage{amssymb}
\usepackage{float}

\title{Comorbidity-Informed Transfer Learning for Neuro-developmental Disorder Diagnosis}

\author[1]{Xin Wen}
\author[1]{Shijie Guo}
\author[1]{Wenbo Ning}
\author[1]{Rui Cao}
\author[2]{Jie Xiang}
\author[3*]{Xiaobo Liu}
\author[4*]{Jintai Chen}

\affil[1]{School of Software, Taiyuan University of Technology, Taiyuan, Shanxi, China}
\affil[2]{School of Computer Science(Data Science),Taiyuan University of Technology, Taiyuan, Shanxi, China}
\affil[3]{Montreal Neurological Institute ,Mcgill University, Quebec, Canada}
\affil[4]{Hong Kong University of Science and Technology’s Guangzhou campus, Guangdong, China}

\begin{document}

\maketitle
\footnote{* Corresponding Author}
\begin{abstract}
Neuro-developmental disorders are manifested as dysfunctions in cognition, communication, behaviour and adaptability, and deep learning-based computer-aided diagnosis (CAD) can alleviate the increasingly strained healthcare resources on neuroimaging. However, neuroimaging such as fMRI contains complex spatio-temporal features, which makes the corresponding representations susceptible to a variety of distractions, thus leading to less effective in CAD. For the first time, we present a Comorbidity-Informed Transfer Learning(CITL) framework for diagnosing neuro-developmental disorders using fMRI. In CITL, a new reinforced representation generation network is proposed, which first combines transfer learning with pseudo-labelling to remove interfering patterns from the temporal domain of fMRI and generates new representations using encoder-decoder architecture. The new representations are then trained in an architecturally simple classification network to obtain CAD model. In particular, the framework fully considers the comorbidity mechanisms of neuro-developmental disorders and effectively integrates them with semi-supervised learning and transfer learning, providing new perspectives on interdisciplinary. Experimental results demonstrate that CITL achieves competitive accuracies of 76.32\% and 73.15\% for detecting autism spectrum disorder and attention deficit hyperactivity disorder, respectively, which outperforms existing related transfer learning work for 7.2\% and 0.5\% respectively.
\end{abstract}
\begin{figure*}[ht]
\begin{center}
\centerline{\includegraphics[width=\textwidth]{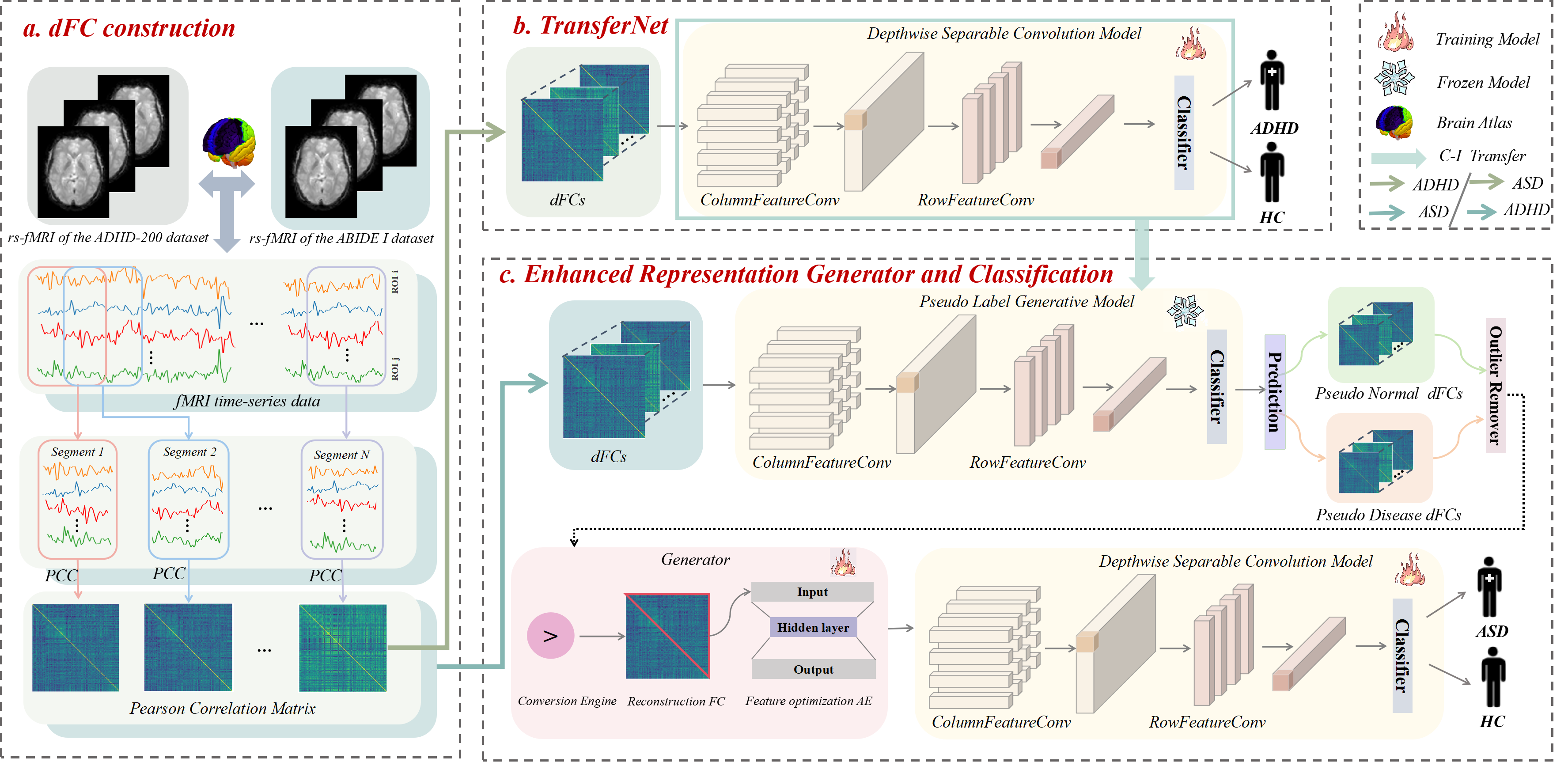}}
\caption{illustrates the proposed CITL framework, which consists of three main components: (a) Construction of dFCs, (b) Pre-trained TransferNet, which is used to screen out the dFCs that possess comorbid patterns, (c) Enhanced Representation Generator and Classification which contains an encoder-decoder generator and an architecturally simple classifier.}
\label{figure1}\centering
\end{center}
\end{figure*}
\section{Introduction}
Neuro-developmental disorders are characterized by impairments in cognition, communication, behavior, and adaptability \cite{thapar2017Neurodevelopmental}, such as Autism Spectrum Disorder (ASD) and Attention Deficit Hyperactivity Disorder (ADHD) \cite{lauritsen2013autism,tripp2009neurobiology}. These disorders have become a significant global public health issue, early diagnosis and intervention are crucial for improving patients' life \cite{world2006neurological}. Deep learning-based computer-aided diagnosis (CAD) has shown tremendous potential in providing efficient, automated diagnostic tools and analyzing medical imaging data \cite{doi2007computer}, which utilize brain functional connectivity (FC) extracted from functional magnetic resonance imaging (fMRI) for disease diagnosis \cite{canario2021review,ingalhalikar2021functional,huang2020multi}. FC, as dynamic interaction information between brain regions, could reveal the neural mechanisms of neuro-developmental disorders and provide potential insights into brain function \cite{du2024survey}. 

In the recent CAD of neuro-developmental disorders, the scarcity of data samples has emerged as a significant bottleneck and transfer learning has been increasingly incorporated into the diagnostic framework which is accompanied by sophisticated domain adaptation techniques\cite{hu2023source,wang2024brainsck}. However, domain adaptation methods\cite{farahani2021brief,shi2021domain} generally require pre-selecting source and target domains, involving partitioning the dataset. Divergent partitioning strategies can give rise to substantial oscillations in the outcomes of CAD. This phenomenon renders the efficacy of transfer learning inconsistent when applied to diverse datasets or under varying experimental conditions. On the other hand, ASD and ADHD share a common genetic basis and exhibit overlapping impairments in social and executive functions\cite{brookman2018characterizing,antshel2019autism}. Existing CAD methods\cite{yin2022semi,ingalhalikar2021functional,huang2020multi} tend to focus on the standalone diagnosis, without comprehensively investigating the comorbid mechanisms between ASD and ADHD\cite{zablotsky2020co,antshel2019autism}, which holds the potential to augment representation learning, thereby facilitating more accurate diagnostic results.

The fMRI signals harbor intricate spatio-temporal information\cite{liu2023braintgl,gadgil2020spatio}. A fundamental hypothesis indicates that during the entire scanning period, the dynamic functional connectivity(dFC) of fMRI exhibits disease-related abnormal characteristics solely at certain specific time points\cite{Karahanolu2015TransientBA,dfc}. Nowadys, dFC is frequently utilized as a data augmentation strategy in CAD\cite{10048009,wang2021modeling}. Nevertheless, this approach contravenes the fundamental postulate of independent and identically distributed (i.i.d.) within the dataset. Such a violation elevates the vulnerability to data leakage, thereby calling into question the credibility of the research outcomes. Consequently, it is necessary to integrate the dFC data of the same subject into a single feature to meet the requirement of i.i.d. data. Semi-supervised learning\cite{hu2023source,mengi2024ssmda,li2023semi} combined with pseudo-labeling\cite{hu2022dual,rifat2024semi,shenaj2023federated} can iteratively identify the required information from a large number of unlabeled samples, showing potential in resolving the aforementioned issues. This also represents a novel attempt from an interdisciplinary perspective.

Therefore, we propose the CITL framework, which leverages the shared comorbid mechanisms between ASD and ADHD within the context of transfer learning. Specifically, TransferNet is initially trained to solely detect ADHD, after which it is frozen and subsequently transferred to analyze the dFCs associated with ASD. The frozen model is then employed to preserve dFCs linked to comorbid mechanisms, using a pseudo-labeling technique in combination with a self-supervised learning approach to iteratively refine the classification of each subject’s dFC. Our hope is to enhance the model’s ability to capture subtle, comorbidity-driven features during the model transfer.
To ensure that the enhanced representation meets the \textit{i.i.d.} condition (for preventing data leakage), a novel generator with an encoder-decoder architecture is proposed. Distinct from conventional generative models, the output of the encoder in our proposed generator serves as the generated enhanced FC. This design can, to a certain degree, alleviate the influence of the low signal-to-noise ratio in fMRI. Under the guidance of the comorbidity mechanism, it enables the generation of enhanced FC with significantly stronger discriminability. Finally, a simple MLP-based classifier is used to detect neuro-developmental disorders. Furthermore, to validate the effectiveness of CITL, additional experiments are conducted, including ablation experiments and other transfer sequences, providing more insights into its capabilities and limitations. The specific innovative points are as follows:

\begin{itemize}
    \item A novel comorbidity informed transfer learning is proposed. Based on the comorbidity mechanism between ASD and ADHD, dFCs from each individual with comorbidity-related are selected for enhanced representation generation. 
    \item A Conversion Engine and enhanced representation generator are proposed. Composed of pseudo-labeling and autoencoder, the enhanced functional connectivity is generated with i.i.d. criterion. Such could effectively mitigate the impact of low signal-to-noise ratio in fMRI on classification tasks. 
    \item The CITL framework has achieved the state-of-the-art performance compared with various transfer learning approaches for diagnosing ASD and ADHD. And the enhanced functional connectivity generated, guided by the comorbidity mechanism, exhibits higher discriminability and also provides a new interpretation of the physiological mechanism.
\end{itemize}

\begin{figure*}[ht]
\centerline{\includegraphics[width=0.7\textwidth]{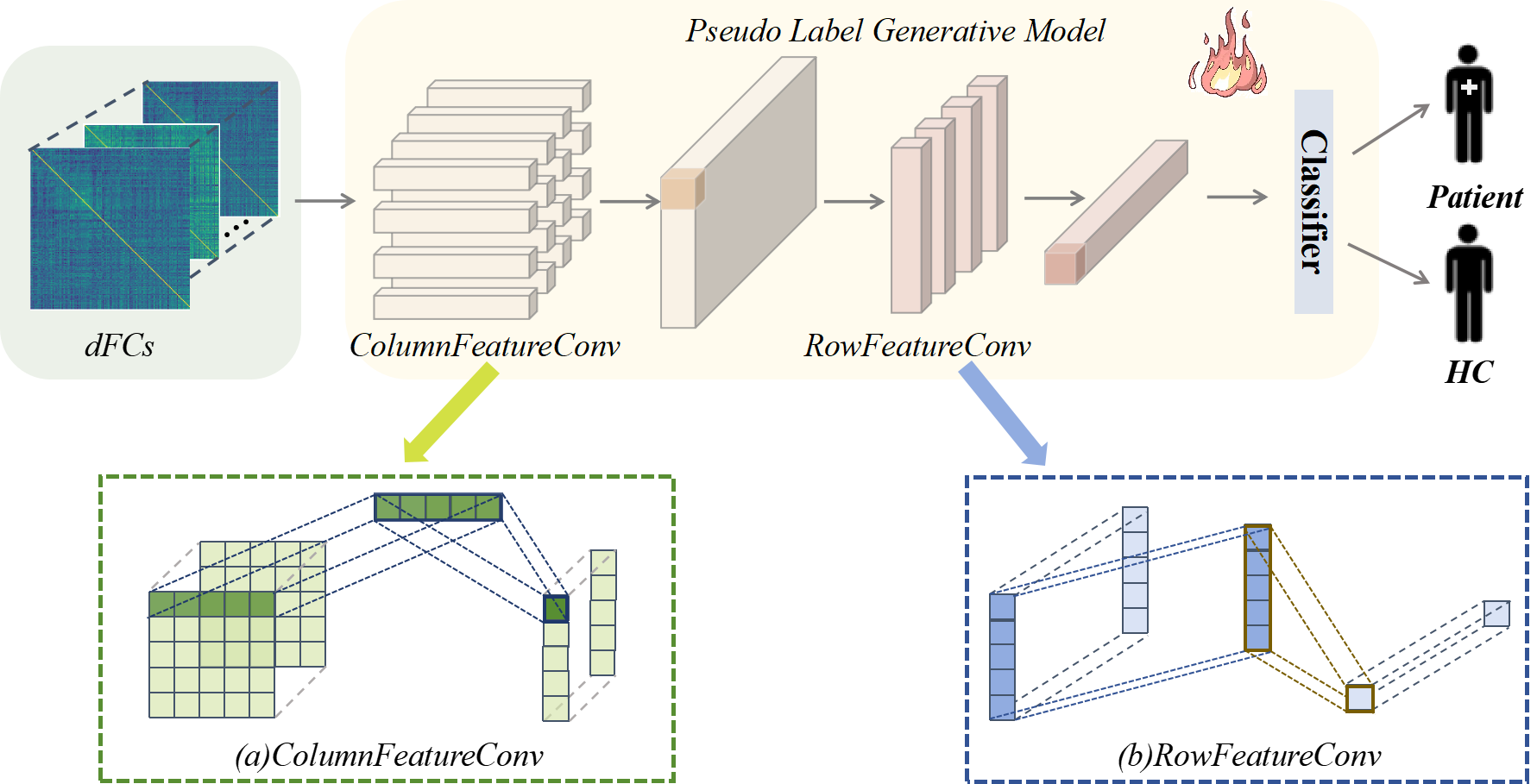}}
\caption{ illustrates the two convolution modules of the Depthwise Separable Convolution Model: (a) represents the ColumnFeatureConv Model, which primarily focuses on edge information, and (b) shows the RowFeatureConv module, which is used to integrate node features.}
\label{figure2}
\end{figure*}
\section{Methodology}
The CITL framework we proposed is shown in Figure 1. First, dFCs are computed separately from the ABIDE I and ADHD-200 datasets(Figure 1a). Then, dFCs from ADHD-200 are fed into the TransferNet module (Figure 1b), where a pre-trained transfer model is obtained through training. The parameters of this model are then frozen and transferred to the Pseudo Label Generative Model(Figure 1c). Subsequently, dFC from the ABIDE I are fed into the pseudo-label generation model, where the dFCs of each participant are assigned with pseudo-labels which reflect the comorbidity mechanism between ASD and ADHD. And an Enhanced Representation Generator is used to filter out "low-correlated dFCs" that are less relevant to comorbidity, retaining the latent features highly associated with comorbidity. Finally, the generated functional connectivity is fed into the Depthwise Separable Convolution Model for classification.
\subsection{dFCs construction}
As depicted in Figure 1a, dFCs of a subject in this study were obtained using sliding window approach\cite{hindriks2016can}. The resting-state fMRI data from the ABIDE dataset and the ADHD-200 dataset were processed with the AAL (116) atlas \cite{rolls2020automated} to extract time series for the 116 predefined brain regions. A sliding window method was applied to these time series to segment them into multiple overlapping windows, capturing their local temporal features. Let the extracted time series be represented as: \(
\mathbf{X} = [\mathbf{x}_1, \mathbf{x}_2, \ldots, \mathbf{x}_T]
\), where \( T \) is the length of the time series, and \( \mathbf{x}_t \in R^{116} \) represents the signals across the 116 regions at time \( t \). For a window size \( W \) and step size \( S \), the \( i \)-th window starting at index \( i \) can be expressed as: \(
\mathbf{W}_i = [\mathbf{x}_i, \mathbf{x}_{i+1}, \ldots, \mathbf{x}_{i+W-1}]
\), The starting index \( i \) begins at 0 and increments by \( S \) with each step until the condition \( i + W - 1 \leq T \) is satisfied.  

 Pearson correlation is employed to calculate dFC for each window in Equation (1):  
\begin{equation}
dFC_{ij} = \frac{\sum_{k=1}^n (x_{k,i} - \bar{x}_i)(x_{k,j} - \bar{x}_j)}{\sqrt{\sum_{k=1}^n (x_{k,i} - \bar{x}_i)^2 \sum_{k=1}^n (x_{k,j} - \bar{x}_j)^2}}
\end{equation}
where \( n \) is the window length, \( x_{k,i} \) and \( x_{k,j} \) are two time series in the \( i \)-th and \( j \)-th brain regions within the window, and \( \bar{x}_i \) and \( \bar{x}_j \) are their respective mean values. 

For a subject with a time series of length \( L \), the total number of dFCs \( N \) can be determined by Equation (2):  
\begin{equation}
N = \left\lfloor \frac{L - W}{S} \right\rfloor + 1
\end{equation}
The set of dFC for a subject is denoted as \( \text{dFC} = \{\mathbf{dFC}_1, \mathbf{dFC}_2, \ldots, \mathbf{dFC}_N\} \).  
\subsection{TransferNet}
The dFCs from one of the datasets are fed into the TransferNet model for training, resulting in a model that can be transferred to the next stage for generating pseudo-labels. As illustrated in Figure 1b, the TransferNet model comprises a Depthwise Separable Convolution Model and an MLP module.

Initially, the dFCs are fed into the ColumnFeatureConv module as described in Equation (3). Pervious approaches focused on neighbor nodes or other local information, while ColumnFeatureConv module aggregates information from edges connected to each node, transforming edge information into node feature representations, resulting in a global perspective. 
\begin{equation}
Y_1[i, j] = \sum_{m=0}^{c_1-1} \sum_{n=0}^{\text{dim}-1} X[i, j+n] \cdot W_1[m, n] + b_1
\end{equation}
where \( X \) is the input feature matrix, \( Y_1 \) is the output matrix, \( W_1 \) is the convolution kernel, \( b_1 \) is the bias term, \( i \) and \( j \) are the indices of the output matrix, \( \text{dim} \) represents the dimension of the feature matrix, and \( c_1 \) is the number of output channels in the first convolution layer.

\begin{equation}
Y_2[i, j] = \sum_{m=0}^{c_2-1} \sum_{n=0}^{\text{dim}-1} \tanh\left(Y_1[i+n, j]\right) \cdot W_2[m, n] + b_2
\end{equation}
where \( W_2 \) is the convolution kernel, \( Y_2 \) is the convolution output, \( b_2 \) is the bias term, \( i \) and \( j \) are the indices of the output matrix, and \( C_2 \) is the number of output channels in the second convolution layer.

The RowFeatureConv module aggregates information from all nodes into a global representation. Subsequently, the data is passed through the hyperbolic tangent activation function (\(\tanh\)) and then flattened into a lower-dimensional latent feature representation.

In summary, within the Depthwise Separable Convolution Model, the dFC first focuses on edge information in the ColumnFeatureConv module and then integrates node features in the RowFeatureConv module. This sequential process enhances the ability to consolidate global information, effectively reducing information loss and improving the efficiency of information utilization. The structure of the Depthwise Separable Convolution Model is illustrated in Figure 2. Finally, the obtained latent features are passed through the MLP module for classification. 

The model is trained using the loss function CrossEntropyLoss\cite{mao2023cross}(Equation 5), and the performance of the model is evaluated based on its classification accuracy. The best-performing transfer model, referred to as the Pseudo Label Generative Model, is then saved for generating pseudo-labels in the subsequent stage. The CrossEntropyLoss is given by:
\begin{equation}
\text{CrossEntropyLoss} = -\frac{1}{N} \sum_{n=1}^N \sum_{i=1}^C y_{n,i} \log(\hat{y}_{n,i})
\end{equation}
where \( N \) is Total number of samples in the dataset. \( C \) is Total number of classes. \( y_{n,i} \)The true label for the \( n \)-th sample and the \( i \)-th class. \( \hat{y}_{n,i} \)is the predicted probability for the \( n \)-th sample belonging to the \( i \)-th class.  
\begin{figure*}
\centerline{\includegraphics[width=\textwidth]{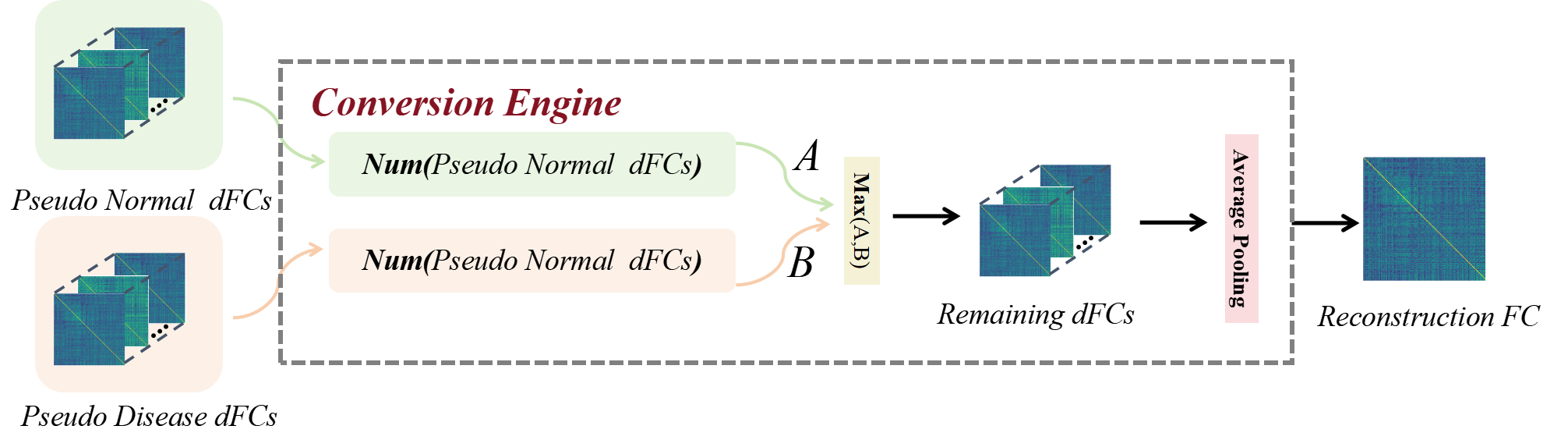}}
\caption{The Conversion Engine removes the set with fewer samples and then optimizes the remaining dFCs through average pooling. The dFCs are transformed into the Reconstruction FC}
\label{figure3}
\end{figure*}
\subsection{Pseudo Label Generative Model and Enhanced Representation Generator}
The dFCs from the second dataset are input into the Pseudo Label Generative Model, which is obtained through transfer learning from the previous stage. The pseudo-label generative model acts as a frozen model. Since each subject corresponds to multiple dFCs, when the dFCs of each subject are predicted by the pseudo-label generative model, two types of pseudo-labels are generated: Normal Pseudo Label and Disease Pseudo Label(Figure 1c). This process is expressed by Equations (6) and (7), the specific method of Pseudo Label Generative is presented in Algorithm 1.
\begin{equation}
\hat{y_i} = Pseudo \ Label \  generative \ Model(X_i))
\end{equation}
\begin{equation}
\text{dFCs}=
\begin{cases}
    Normal\ Pseudo\ Label\ dFCs\quad \text{if }\ \hat{y_i}=0\\
    Disease\ Pseudo\ Label\ dFCs\quad \text{if }\ \hat{y_i}=1
\end{cases}
\end{equation}
where the Pseudo Label Generative Model is the best model obtained from the previous stage and applied through transfer learning, where \( X \) represents the dFCs.

A subject's dFC  calculated over different time intervals may include data that are not strongly correlated with disease diagnosis, which can interfere with downstream tasks such as prediction classification and hinder achieving optimal results. To better extract the subject's features, we propose the Enhanced Representation Generator. It consists of two parts: the Conversion Engine and the Feature Optimization model. First, the Conversion Engine, evaluates two sets of pseudo-labeled dFCs obtained from the subject(Figure 3). It removes the set with fewer samples and then optimizes the remaining dFCs through average pooling. After passing through the Conversion Engine, the dFCs are transformed into the Reconstruction FC,(Equation 8).
\begin{equation}
\mathbf{X}_{\text{Reconstruction}} = 
\begin{cases} 
\frac{1}{N_1} \sum_{i=1}^{N_1} X_{1,i}, & \text{if } N_1 \geq N_2 \\[10pt]
\frac{1}{N_2} \sum_{i=1}^{N_2} X_{2,i}, & \text{if } N_2 > N_1
\end{cases}
\end{equation}
here, \( X_1 \) represents the set of dFCs with pseudo-label \( y=1 \), containing \( N_1 \) samples, while \( X_2 \) represents the set of dFCs with pseudo-label \( y=0 \), containing \( N_2 \) samples.
\begin{algorithm}[tb]
    \caption{Pseudo Label Generative}
    \label{alg:algorithm}
    \textbf{Input:} Training set \(X_1\), Testing set \(X_2\), Pseudo Label generative set \(X_3\). The training set \(X_1\) and the testing set \(X_2\) consist of dFCs from the first dataset, split in an 80:20 ratio, while \(X_3\) contains dFCs from the second dataset.\\
    \textbf{Output:} The second dataset contains dFCs with pseudo-labels.
    \begin{algorithmic}[1]
        \STATE \textbf{Initialization:} loss, weights, \(i\) and biases
        \WHILE {Epoch \(> i\)}
        \STATE \textbf{Forward propagation:}\\
        \STATE \hspace{2em} \( y = \text{Depthwise Separable Convolution}(X_1)\)
        \STATE \textbf{Calculate the loss:}\\
        \STATE \hspace{8em} \( \text{loss} = \text{CrossEntropyLoss}(y) \)
        \STATE \textbf{Back-propagation:}\\
        \STATE \hspace{8em} \( \arg \min_y \text{loss} + \lambda \|W\|^2 \)
        \STATE \( i \gets i + 1 \)
        \STATE Save(Model)
        \ENDWHILE
        \STATE\textbf{Save Pseudo Label Generative Model:}\\
        \STATE\hspace{11em}Save(\( Model_{BestAcc}\) )\\
        \STATE\textbf{Pseudo Label Generative:}\\
        \STATE\hspace{4em}\(\hat{y}\) = \text{Pseudo Label Generative Model}\((X_3)\)\\
        \STATE\hspace{1em}\textbf{if} {\(\hat{y} == 0\)} \textbf{then}\\
        \STATE\hspace{2em}\(Data_{Pseudo Label}\) = Normal Pseudo Label dFCs\\
        \STATE\hspace{1em}\textbf{else}\\
        \STATE\hspace{2em}\(Data_{Pseudo Label}\) = Disease Pseudo Label dFCs\\
        \STATE\hspace{1em}\textbf{end if}\\
        \STATE\textbf{Save Pseudo Label and Data}
\end{algorithmic}
\end{algorithm}

Then, the Reconstruction FC is fed into the Feature Optimization Auto-encoder(AE) model(Figure 4, Equation 9)
\begin{equation}
V = \{ \text{Reconstruction } FC_{ij} \mid 1 \leq i < j < n \}    
\end{equation}
The Feature Optimization AE model which is composed of two parts: encoder and decoder. The encoder encodes the input data into a low-dimensional representation(Equation 10). The decoder reconstructs the output of the encoder back to the original input data(Equation 11). Finally, the variables from the hidden layer are output and transformed into the upper triangular part of a matrix. The upper triangular matrix is then transposed, and the diagonal elements are assigned a value of 1, resulting in a symmetric matrix referred to as the Optimization FC.
\begin{equation}
 Z_{\text{encoder}} = f_{\text{encoder}}(V) = \sigma \left( V W_{\text{encoder}} + b_{\text{encoder}} \right)   
\end{equation}
\begin{equation}
X^{\prime}=f_{\text {decoder }}\left(Z_{\text {encoder }}\right)=Z_{\text {encoder }} W_{\text {decoder }}+b_{\text {decoder }}  
\end{equation}
where \(\sigma\) is the activation function, \(W_{encoder}\) and \(b_{encoder}\) are the weight and bias of the encoder, respectively. 
\begin{figure*}
\centerline{\includegraphics[width=\textwidth]{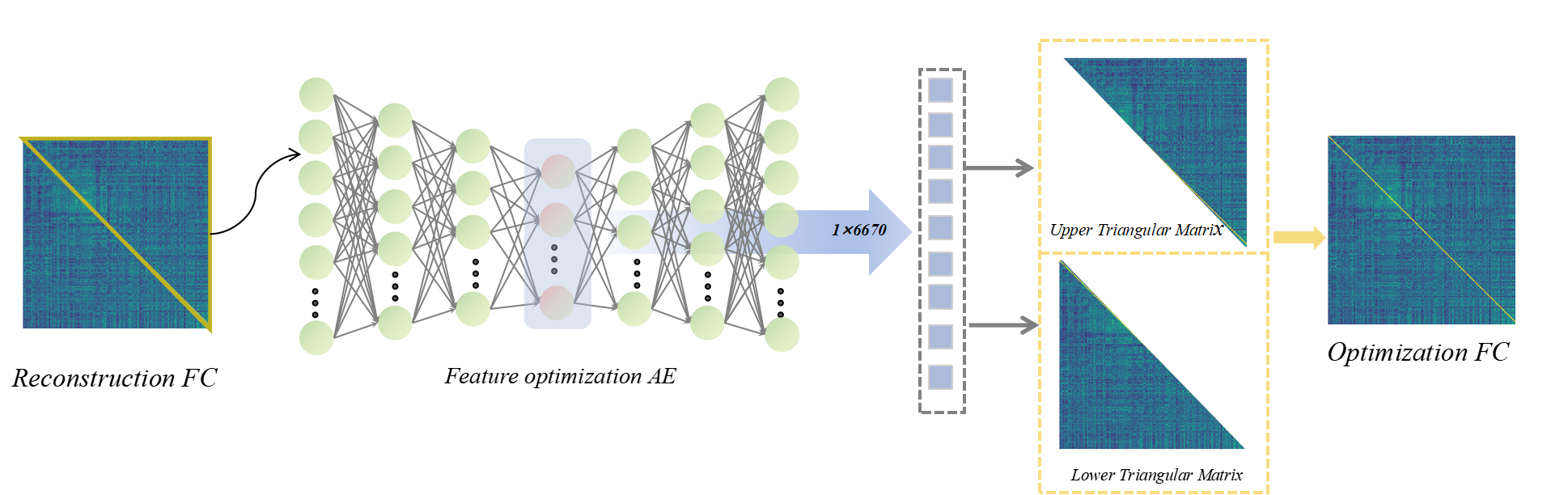}}
\caption{The Feature Optimization AE model. The variables from the hidden layer are transformed into a 6670-dimensional vector, which represents the upper triangular part of the 116 templates and is used to construct the upper triangular part of the matrix. This matrix is transposed, and its diagonal elements are set to 1, resulting in a symmetric matrix referred to as the Optimization FC.}
\label{figure4}
\end{figure*}
The original dFCs are processed through the Pseudo Label Generative Model and Enhanced Representation Generator, which removes data with low correlation to disease diagnosis(low-correlated dFCs), while obtaining the hidden representations of the data.
\subsection{Classification}
The extracted Optimization FC is classified using the Depthwise Separable Convolution Model. The classifier is trained jointly with the Optimization FC, two loss functions are set for training. The loss function for training the Depthwise Separable Convolution Model is Cross-Entropy loss(Equation 5), while the loss function for training the Feature Optimization AE model is the Cosine Embedding Loss(Equation 12). 
\begin{equation}
L\left(x_1, x_2, y\right)= \begin{cases}1-\cos \left(x_1, x_2\right) & \text { if } y=1 \\ \max \left(0, \cos \left(x_1, x_2\right)+1\right) & \text { if } y=-1\end{cases}
\end{equation}

The value of \( y \) is 1, indicating that a high similarity is desired between \( x_1 \) and \( x_2 \), and -1, indicating that a low similarity is desired. In this paper, we aim for a high similarity between the Reconstruction FC and the Optimization FC. \(\cos \left(x_1, x_2\right)\) is the cosine similarity between \( x_1 \) and \( x_2 \), and the calculation is given by Equation (13).
\begin{equation}
\cos \left(x_1, x_2\right)=\frac{x_1 \cdot x_2}{\left\|x_1\right\|\left\|x_2\right\|}
\end{equation}
\subsection{Implementation details}
CITL is implemented in PyTorch\cite{imambi2021pytorch}. The experiments are accelerated on an NVIDIA GeForce RTX 3090 24G. Hyperparameter tuning is performed using the NNI automated tuning tool \cite{pythonautomated}, which helps optimize the model's performance by automatically adjusting parameters. The final implementation details are as follows: the learning rate \( l_1 \) for the Enhanced Representation Generator model is set to 0.00008, and the weight decay coefficient \( w_1 \) is set to 0.0005. For the Depthwise Separable Convolution Model, the learning rate \( l_2 \) is set to 0.00004, and the weight decay coefficient \( w_1 \) is set to 0.0005. The batch size is set to 64.
\section{Results}
\subsection{Datasets and Preprocessing}
ABIDE I, II\cite{di2014autism} and ADHD-200\cite{adhd2012adhd} datasets are involved to evaluate CITL. In this work, we used 1,035 participants from the ABIDE I dataset (505 ASD and 530 NC samples), 586 participants from the ABIDE II dataset (264 ASD and 322 NC samples), and 782 participants from the ADHD-200 dataset (353 ADHD and 429 NC samples).The number of subjects included in this study and their demographics are given in Table 1.
\begin{table}[H]
\centering
\resizebox{\columnwidth}{!}{
\begin{tabular}{lcccccc}
\toprule
\textbf{Dataset} & \multicolumn{2}{c}{\textbf{ABIDE I}} & \multicolumn{2}{c}{\textbf{ABIDE II}} & \multicolumn{2}{c}{\textbf{ADHD-200}} \\
\cmidrule(lr){2-3} \cmidrule(lr){4-5} \cmidrule(lr){6-7}
 \textbf{Information} & \textbf{ASD} & \textbf{NC} & \textbf{ASD} & \textbf{NC} & \textbf{ADHD} & \textbf{NC} \\
\midrule
\textbf{Subjects} & 505 & 530 & 264 & 322 & 353 & 429 \\
\textbf{Gender (F/M)} & 62/443 & 95/435 & 44/231 & 109/226 & 75/277 & 207/222 \\
\textbf{Age (mean ± std)} & 17.1±8.6 & 16.8±7.5 & 13.9±8.0 & 13.5±7.5 & 11.6±3.0 & 11.7±3.2 \\
\bottomrule
\end{tabular}
}
\caption{Demographic information of studied subjects in ABIDE I, ABIDE II and ADHD-200. ASD: autism spectrum disorder, NC: normal control, ADHD: attention deficit hyperactivity disorder.}
\end{table}
The preprocessed connectome project (PCP) \cite{craddock2013neuro} was used to preprocess the ABIDE dataset. The preprocessing pipeline applies a configurable pipeline for analysis of connectomes (CPAC). and Athena pipeline \cite{bellec2017neuro} is used for preprocessing ADHD-200 dataset. The preprocessing work was completed by Cameron Craddock at the Athena computer cluster of Virginia Institute of Technology.
\begin{table*}[ht]
\centering
\begin{tabular}{llccccc}
\toprule
\textbf{Dataset} & \textbf{Method} & \textbf{Subjects} & \textbf{ACC} & \textbf{SEN} & \textbf{SPE} & \textbf{AUC} \\
\midrule
\multirow{9}{*}{ABIDE I} 
    & BrainGNN\cite{li2021braingnn}  & 1035  & 69.31 ± 3.03 & 68.18 ± 11.05& 68.64 ± 13.77 & 69.52 ± 3.52 \\
    & BrainNetCNN\cite{kawahara2017brainnetcnn} & 1035 & 69.73 ± 2.76 & 66.73 ± 9.93 & 72.54 ± 8.26  & 71.99 ± 3.06 \\
    & BrainSCK(TL)\cite{wang2024brainsck} & 819   & 63.6 & -& - & -  \\
    & FMTLJD(TL) \cite{huang2022federated} &1035 & 69.48 ± 1.6 & 73.38 ± 4.8 & 69.65 ± 3.8  & - \\
    & S3TL(TL)\cite{hu2023source}       &1035  & 69.14 ± 3.87 & 67.27 ± 4.32 & 70.88 ± 3.21  & 72.52 ± 3.56 \\
    & Ours(TL)       &1035  & \textbf{76.32 ± 3.36} & \textbf{75.96 ± 6.10} & \textbf{77.46 ± 6.66} & \textbf{76.24 ± 3.57} \\
\midrule
\multirow{9}{*}{ADHD-200} 
    & BrainGNN\cite{li2021braingnn}   & 782 & 66.63 ± 5.41 & 70.10 ± 12.77& 63.97 ± 10.79 & 64.03 ± 3.88  \\
    & BrainNetCNN\cite{kawahara2017brainnetcnn} & 782& 63.77 ± 3.84 & 69.87 ± 12.76& 58.37 ± 13.57 & 62.97 ± 5.10  \\
    & BrainSCK(TL)\cite{wang2024brainsck} & 766    & 70.1 & -& - & -  \\
    & FMTLJD(TL)\cite{huang2022federated}  &939 & 71.44 ± 3.2 & 65.55 ± 6.3&75.68 ± 2.8 & -  \\
    & S3TL(TL)\cite{hu2023source}     &939    & 72.62 ± 3.96 & 62.71 ± 3.65 & \textbf{77.43 ± 4.32}  & \textbf{73.66 ± 2.85}  \\
    & Ours(TL)     &  782  & \textbf{73.15 ± 3.73} & \textbf{72.95 ± 3.99} & 73.27 ± 5.48 & 72.55 ± 3.57 \\
\bottomrule
\end{tabular}%
\caption{The comparison with the existing methods for ABIDE I and ADHD-200 Datasets. The results (in \%) were calculated based on 10-fold cross-validation. The best result in each category is highlighted in bold. (TL) indicates that the method utilizes transfer learning.}
\end{table*}
\begin{figure}
\centerline{\includegraphics[width=0.9\columnwidth]{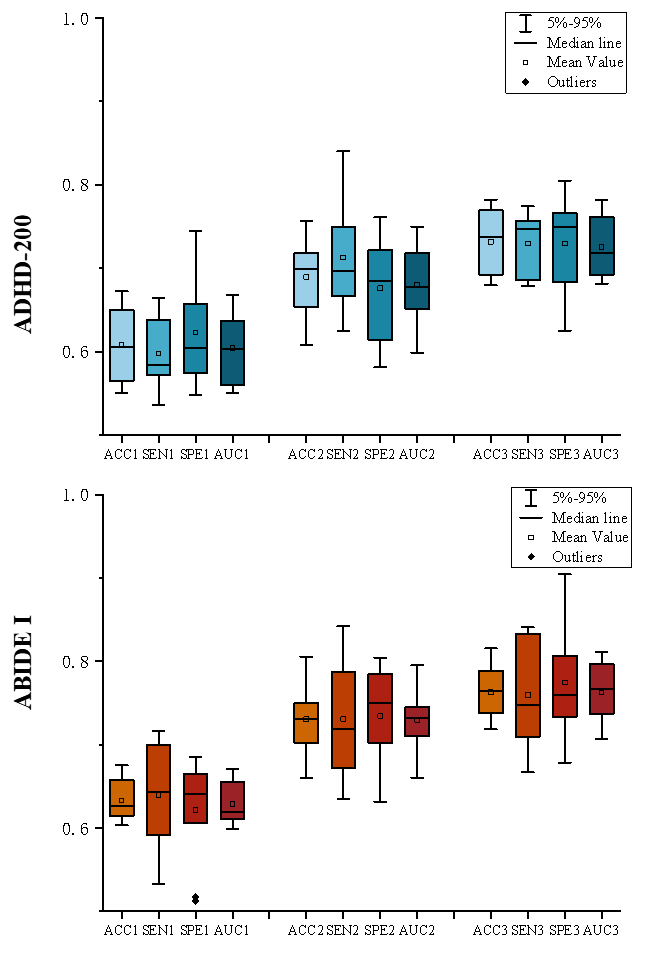}}
\caption{The ablation experiments of different modules. The result is shown in mean±std over 10-fold cross validation. On the x-axis, '1' represents the results of ablation experiment (1), '2' represents the results of ablation experiment (2) and '3' represents the results of ablation experiment (3).}
\label{figure5}
\end{figure}
\subsection{Competing methods}
To evaluate the proposed CITL framework, we compare it with typical classification frameworks for brain data, such as BrainGNN \cite{li2021braingnn} and BrainNetCNN \cite{kawahara2017brainnetcnn}. we also compare it with recently proposed methods that employ transfer learning for brain data classification, such as: BrainSCK\cite{wang2024brainsck}, FMTLJD \cite{huang2022federated}, S3TL\cite{hu2023source}.

\begin{table*}[ht]
\centering
\begin{tabular}{@{}>{\centering\arraybackslash}m{2.5cm} 
                >{\centering\arraybackslash}m{2.5cm} 
                >{\centering\arraybackslash}m{1.5cm} 
                >{\centering\arraybackslash}m{2cm} 
                >{\centering\arraybackslash}m{2cm} 
                >{\centering\arraybackslash}m{2cm} 
                >{\centering\arraybackslash}m{2cm}@{}}
\toprule
\textbf{Dataset} & \textbf{TL and Pseudo Label Generative} & \textbf{Enhanced Representation Generator} & \textbf{ACC} & \textbf{SEN} & \textbf{SPE} & \textbf{AUC} \\
\midrule
\multirow{3}{*}{ADHD-200$\rightarrow$ABIDE I} 
 & x & x & 63.30$\pm$2.65 & 63.93$\pm$6.84 & 62.14$\pm$6.11 & 64.88$\pm$2.68 \\
 & \checkmark & x & 73.04$\pm$4.03 & 73.06$\pm$7.25 & 73.44$\pm$5.58 & 74.97$\pm$3.71 \\
 & \checkmark & \checkmark & \textbf{76.32$\pm$3.36} & \textbf{75.96$\pm$6.10} & \textbf{77.46$\pm$6.66} & \textbf{76.24$\pm$3.57} \\
\midrule
\multirow{3}{*}{ABIDE I$\rightarrow$ADHD-200} 
 & x & x & 60.85$\pm$4.45 & 58.99$\pm$3.87 & 62.27$\pm$6.98 & 61.36$\pm$4.28 \\
 & \checkmark & x & 68.93$\pm$4.36 & 71.26$\pm$6.35 & 67.60$\pm$6.54 & 69.02$\pm$4.52 \\
 & \checkmark & \checkmark & \textbf{73.15$\pm$3.73} & \textbf{72.95$\pm$3.99} & \textbf{73.27$\pm$5.48} & \textbf{72.55$\pm$3.57} \\
\bottomrule
\end{tabular}
\caption{Ablation results for CITL Model on ABIDE I and ADHD-200 datasets.}
\end{table*}

\subsection{Classification Performance}
This work compares the classification performance of different machine learning and deep learning methods on the ABIDE I and ADHD-200 datasets. The proposed method achieves accuracy of 76.32\% and 73.15\%, respectively. The proposed model also shows sensitivity and specificity of 75.96\% and 77.46\% on the ABIDE I dataset, and 72.95\% and 73.27\% on the ADHD-200 dataset. Compared to other methods, the results are more balanced, indicating that CITL has a well-balanced predictive ability and can accurately distinguish between patients and controls.
\subsection{Ablation studies}
The main advantage of this model lies in its ability to extract latent features of functional connectivity (FC) through the mutual transfer learning between the ABIDE I and ADHD-200 datasets, combined with the Pseudo Label Generative and Enhanced Representation Generator modules. The Pseudo Label Generative Model and Enhanced Representation Generator modules are used sequentially, meaning that the output of the Pseudo Label Generative Model serves as the input to the Enhanced Representation Generator. To further validate the actual contribution of these components, ablation experiments were conducted. All experiments were performed using 10-fold cross-validation, and the results were evaluated using accuracy, sensitivity, specificity, and AUC. When using transfer learning and the Pseudo Label Generative model, the accuracy improves by approximately 10\% (ABIDE I) and 8\% (ADHD-200) compared to directly classifying each dataset individually. Further inclusion of the Enhanced Representation Generator module leads to an additional increase in accuracy by 3\% and 4\%, respectively(Figure 5).

\textbf{(1) The impact of joint use of ABIDE I and ADHD-200: }To optimize the feature data of the datasets, the joint use of ABIDE I and ADHD-200 serves as the foundation for the Pseudo Label Generative Model and Enhanced Representation Generator. An ablation experiment was conducted by using the Depthwise Separable Convolution Model to classify the two datasets separately. The accuracy for ABIDE I is 63.3\%, and the accuracy for ADHD-200 is 60.85\%(Table 3).

\textbf{(2)The impact of the Pseudo Label Generative Model: }The pseudo-labels are generated by jointly using the ABIDE I and ADHD-200 datasets, followed by classification using the Depthwise Separable Convolution Model. The accuracy for ABIDE I is 73.04\%, and the accuracy for ADHD-200 is 68.93\%, representing a decrease of 3.28\% and 4.22\%, respectively(Table 3).

\textbf{(3) The impact of the Enhanced Representation Generator module: }To further optimize the features, the Enhanced Representation Generator is employed. To assess its impact on the model, this module is added to the previous modules, and the dataset is classified using the complete workflow of this paper. The accuracy for ABIDE I is 76.32\%, and the accuracy for ADHD-200 is 73.15\%(Table 3). Both sensitivity and specificity are well-balanced, indicating that the Enhanced Representation Generator module plays a positive role in the proposed model.

\section{Conclusion}
This work introduces a novel CITL framework for diagnosing neuro-developmental disorders using FC. The key innovation lies in leveraging the comorbidity relationship between ADHD and ASD to propose a cross-diagnostic approach that utilizes pseudo-labeling to exclude low comorbidity correlation dynamic functional connectivity. The combination of an autoencoder and the depthwise separable convolution kernels effectively optimizes functional connectivity for classification tasks. Experimental results on the ABIDE-I and ADHD-200 datasets demonstrate the potential of CITL in identifying neuropsychiatric disorders.
\newpage

\bibliographystyle{unsrt}  
\bibliography{references}

\end{document}